\documentclass[review]{elsarticle}

\usepackage{color}
\usepackage{amsmath}
\usepackage{booktabs}
\usepackage{multirow}  
\usepackage{setspace}  
\usepackage{subfigure} 
\usepackage{algorithm} 
\usepackage{algorithmic}

\journal{Journal of \LaTeX\ Templates}

\begin{document}

\begin{frontmatter}

\title{Label Mask for Multi-Label Text Classification}


\author[a1]{Rui Song}
\ead{songrui20@mails.jlu.edu.cn}

\author[a3]{Xingbing Chen}
\ead{1276402580@qq.com}

\author[a4]{Zelong Liu}
\ead{18943698576@163.com}

\author[a4]{Haining An}
\ead{anhn2418@jlu.edu.cn}

\author[a5]{Zhiqi Zhang}
\ead{1005359144@qq.com}

\author[a2]{Xiaoguang Wang\corref{mycorrespondingauthor}}
\cortext[mycorrespondingauthor]{Corresponding author}
\ead{wangxiaog@jlu.edu.cn}

\author[a3]{Hao Xu}
\ead{xuhao@jlu.edu.cn}

\address[a1]{School of Artificial Intelligence, Jilin University}
\address[a2]{Public Computer Education and Research Center, Jilin University}
\address[a3]{College of Computer Science and Technology, Key Laboratory of Symbolic Computing and Knowledge Engineering of Ministry of Education, Jilin University}
\address[a3]{College of Electronic Science and Engineering, Jilin University}
\address[a4]{College of Construction Engineering, Jilin University}
\address[a5]{College of Sotfware, Jilin University}

\begin{abstract}
One of the key problems in multi-label text classification is how to take advantage of the correlation among labels. However, it is very challenging to directly model the correlations among labels in a complex and unknown label space. In this paper, we propose a Label Mask multi-label text classification model (LM-MTC), which is inspired by the idea of cloze questions of language model. LM-MTC is able to capture implicit relationships among labels through the powerful ability of pre-train language models. On the basis, we assign a different token to each potential label, and randomly mask the token with a certain probability to build a label based Masked Language Model (MLM). We train the MTC and MLM together, further improving the generalization ability of the model. A large number of experiments on multiple datasets demonstrate the effectiveness of our method.
\end{abstract}


\end{frontmatter}

\section{Introduction}
Text categorization is a basic task in natural language processing, Multi-label text classification assigns multiple different labels to a document, rather than a document corresponding to a single category. In recent years, multi-label text classification has been widely used in sentiment analysis~\cite{cambria2014senticnet}, topic classification~\cite{yang2016hierarchical}, information retrieval~\cite{gopal2010multilabel}, label recommendation~\cite{katakis2008multilabel}. How to fully capture semantic patterns from original documents, how to extract discriminant information related to corresponding labels from each document, and how to accurately mine the correlation between labels are the three aspects that researchers pay attention to~\cite{xiao2019labelspecific}. Among them, the relevance modeling of complex and unknown label systems has always been the focus of scholars' efforts.

\begin{figure}[h]
	\begin{center}
		\includegraphics[width=0.65\textwidth]{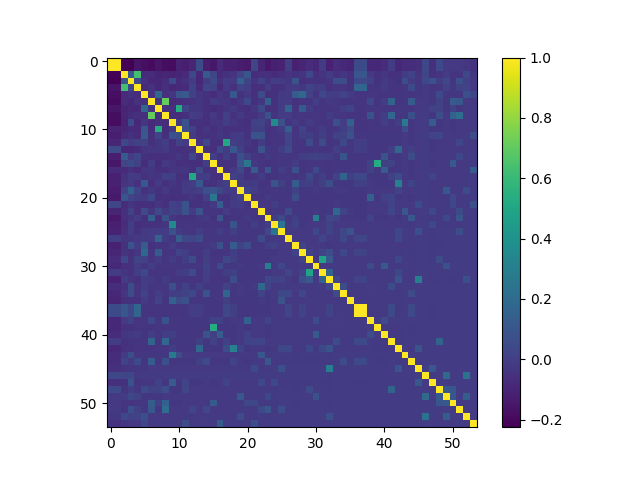}
		\caption{Pearson correlation coefficient between all label pairs in AAPD dataset. The lighter the color, the more relevant the label pairs are.}
		\label{fig:relation}
	\end{center}
\end{figure}

One of the most direct methods to solve multi-label text classification is to transform the multi-label text classification task into several binary classification tasks~\cite{boutell2004learning}, however, this tends to ignore the relationship between multiple labels. Similarly, some deep learning approaches, such as CNN~\cite{liu2017deep} and attention mechanism~\cite{yang2016hierarchical}, can model the documents effectively, but they still ignores the relationship between labels. As shown in Figure~\ref{fig:relation}, there are specific correlations between different label pairs in the dataset AAPD calculated by Pearson correlation coefficient. For labels 0 and 1, the correlation is 1, which means that both appear together in all instances. Therefore, for some datasets with less label information or severe long-tail distribution, the association between labels can provide more important information~\cite{xiao2021does}. 

The emergence of large-scale pre-trained language models, such as Bert  (Bidirectional Encoder Representations from Transformers)~\cite{devlin2018bert} have made knowledge transfer in the field of natural language processing easier. Studies have confirmed that the intermediate layer of Bert encodes a wealth of linguistic information~\cite{jawahar2019what}. Inspired by Cloze Questions (CQ) methods based large-scale pre-trained language models, we propose a Label Mask multi-label text classification model (LM-MTC) to capture the potential semantic and association relations among labels~\cite{schick2020its, schick2021exploiting}. Specifically, we map different labels to different tokens and build a set of token prefix templates. During training, we spliced the token template with the sentences to be classified and input them to Bert. When predicting, we mask all the label tokens and predict them. The advantages of Bert can help the model adaptively capture the semantic relationship between labels and documents, as well as the association relationships. In addition, to make better use of the predictive ability of Bert, we constructed a multi-task framework, randomly masked label tokens, and used the Mask Language Model (MLM) to predict masked tokens to assist in optimizing the multi-label text classification learning task. Our contributions are as follows:
\begin{itemize}
	\item We propose a Label Mask Multi-label Text Classification model (LM-MTC), which converts multi-label text classification into a Cloze Questions (CQ) task and captures the potential relationship between labels with the help of a pre-trained language model. We introduced MLM with LM-MTC for joint training which further improved the performance of the model.
	
	\item Through attention analysis, we confirm the LM-MTC's ability to capture potential associations between labels.
	
	\item We carry out lots of experiments on different types of multi-label text classification tasks to prove the effectiveness of the proposed model.
\end{itemize}

\section{Related Work}
\subsection{Multi-label Text Classification}
Multi-label text classification is a basic task in NLP. There are some methods to solve it by transforming the multi-label text classification task into several binary classification tasks~\cite{boutell2004learning,liu2017deep,yang2016hierarchical}
Some approaches take advantage of pair associations or mutexes between labels. Pairwise comparison (RPC) induces a binary preference relation using a natural extension of pairwise classification, which transforms the multi-label learning task into a label ranking task~\cite{hullermeier2008label}.

However, it is more efficient to assume that a label can be related to multiple labels and take advantage of the higher-order dependencies of the label. Classifier chains (CC) transforms the task of MTC into a chain of binary classification tasks~\cite{read2011classifier}. K-labelsets (RAkEL) constructs small random subsets of labels and transforms MTC into a single-label classification task of random subsets~\cite{tsoumakas2007random}. With the development of deep learning in recent years, some studies have resorted to sequence learning models to solve MTC such as ~\cite{chen2017ensemble}, and Sequence Generation Model (SGM)~\cite{yang2018sgm}. They generate a possible label sequence through the RNN decoder. However, sequence model requires searching for the optimal solution in the potential space, which is too time-consuming when there are too many labels. 

Some approaches model the joint probability distribution of labels, rather than associations for specific labels, such as Bayesian networks~\cite{barutcuoglu2006hierarchical, zhang2010multi-label} and undirected graph models~\cite{wang2015multiple}. \cite{wang2020capturing} strengthen the similarity between the joint distribution of multi-labels and the predicted multi-labels by means of adverse learning framework. In recent years, due to the effectiveness of Graph Neural Network (GNN)~\cite{scarselli2009the} in modeling non-Euclidean spatial data, some methods use GNN to capture the correlation of labels. Label-Specific Attention Network (LSAN) proposes a Label Attention Network model that considers both document content and Label text, and uses self-attention mechanism to measure the contribution of each word to each Label~\cite{xiao2019label-specific}. MAGNET uses a feature matrix and a correlation matrix to capture and explore key dependencies between labels~\cite{ankit2020multi-label}. 

\subsection{Cloze Questions}
A natural way to gain knowledge from a pre-trained MLM is to set the task as fill-in-the-blanks. AUTOPROMPT creates prompts for a diverse set of tasks automatically and shows the inherent ability of MLM to perform emotion analysis and natural language reasoning~\cite{shin2020autoprompt}.~\cite{schick2020its} converts the input text into a cloze problem containing the task description  combined with gradient-based optimization and proves that 'green' LM like Bert can still have competitive performance compared with GPT-3~\cite{b.2020language}. Furthermore, PTE has successfully solved the problems of text classification and natural language reasoning in small samples by using cloze questions method~\cite{schick2021exploiting}. In addition,~\cite{liu2021gpt} propose a similar P-tuning method to improve GPT's ability in natural language understanding tasks.

\section{Preliminaries}

First, we give some notations and describe the MTC task. For the given label space $\mathbf{L}=\{l_1,l_2,...,l_L\}$ and a text $x=\{w_1,w_2,...,w_m\}$, the MTC task aims to learn a a mapping function $\chi:x\to \hat{L}$, where $\hat{L}\subseteq L$ and $|\hat{L}|\ge 1$. Unlike single-label tasks, a text can belong to several different categories. For ease of calculation, the output space of the labels is defined as a vector $L_O\in \mathbf{R}^{|L|}$. For example, when $L=3$ and $\hat{L}=\{l_1,l_2\}$, then $L_O$ is $[1,1,0]$. Then we can rewrite the objective function as $\chi':x\to L_O$. In addition, we define 1 as a positive label and 0 as a negative label.

Given a movie emotion dichotomy sentence, "\textit{The movie was so touching!}", CQ usually generates a new sentence for input by a prefix/suffix template $\tau$: "\textit{The movie was so touching! \textbf{I $\sharp$ it!}}". $\sharp$ can be either "\textit{like}" or "\textit{hate}", indicating positive or negative emotions, respectively. The new input with a prefix template can be expressed as:
\begin{equation}
	\label{eq:x}
	x' = \tau||x
\end{equation}
where $||$ denotes concatenation. 

Consider an MLM $M$ with a vocabulary $V$ and a sentence $x$ with mask $m$, the task of $M$ is to predict the probability of the original word in the mask $p_M^m(w|x)$ where $w\in V$. The goal of CQ is to predict the true value of $\sharp$ by MLM, that is $p_M^m(\sharp |x')$. So the classification task can be transformed into the prediction task of MLM with the help of templates.

\section{LM-MTC}
In this section, we describe the proposed model in detail. The specific execution process is shown in Algorithm~\ref{alg:Framwork}.
\begin{figure*}[h]
	\begin{center}
		\includegraphics[width=0.9\textwidth]{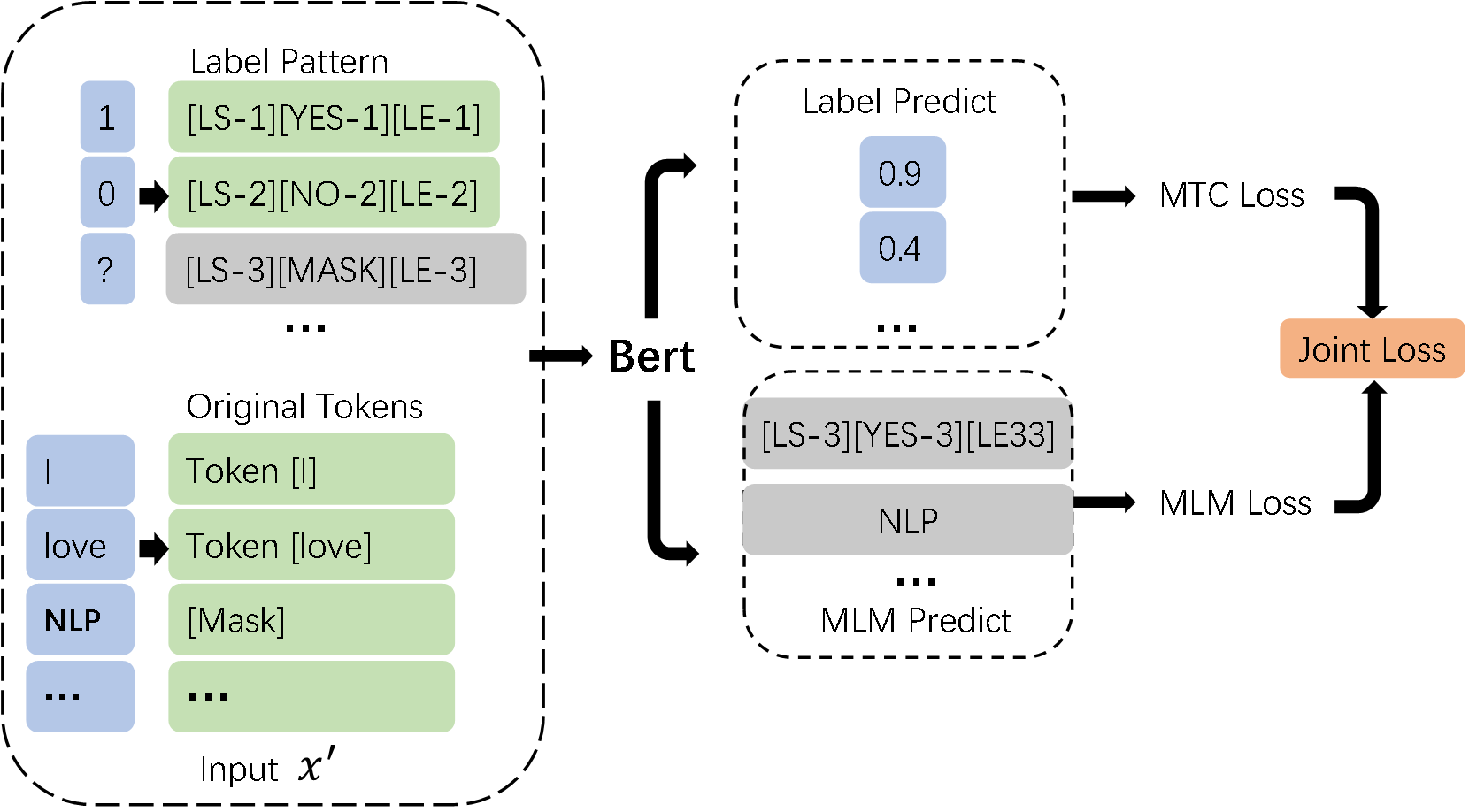}
		\caption{LM-MTC model structure.}
		\label{fig:model}
	\end{center}
\end{figure*}

\subsection{Cloze Patterns}
For cloze tasks, although some studies have demonstrated the advantages of a template approach, it is not clear whether the same template will work for every model, or what kind of template will fit the model better~\cite{shin2020autoprompt}. For MTC, because different documents contain different numbers of true labels and the output is mapped to a single token, it is not possible to build a specific template for each label. To this end, we build a template system for the entire label space. First of all, considering the different states of labels, the labels in each position should have three different states: 0,1 and $mask$. We emphasize the order of the different labels, which is very important for label prediction~\cite{yang2018sgm}. In addition, we also introduce a location-based prompt to allow Bert to clearly recognize where the current Label is located. For label sequence with mask $[1,0,mask]$, we generate the template as follows:
\begin{equation}
	\label{eq:t}
	\begin{split}
		\color[RGB]{0,0,0}{[LS-1][YES-1][LE-1]} \\
		\color[RGB]{0,0,0}{[LS-2][NO-2][LE-2]} \\
		\color[RGB]{0,0,0}{[LS-3][MASK-3][LE-3]}
	\end{split}
\end{equation}
where $LS$ denotes \textit{label start} and $LE$ denotes \textit{label end}.

\subsection{Training and Inference}
\textbf{The forward propagation}. After the template is generated, we treat it as a prefix to the original sentence and input $x'$ to the pre-training model together. The training process has two main objectives: to predict the probability distribution of multiple labels in the label space, and to predict the \textit{mask} by MLM. Assume that the output of Bert is $O\in \mathbf{R}^{|x'|_{max}*768}$, then the prediction of the distribution of the labels and the prediction of the mask can be obtained by using one layer of full connection:
\begin{equation}
	\begin{split}
		O_l = O*W_l+b_l\\
		O_m = O*W_m+b_m
	\end{split}
\end{equation}
where $|x'|_{max}$ denotes max token length, $W_l\in \mathbf{R}^{768*|L|}$. In order for MLM to predict the mask, we need to extend $V$ to $V'$. $V'$ depends on the size of the label space, so that $W_m\in \mathbf{R}^{768*|V'|}$. 

\textbf{Joint loss}. We use the Binary Cross Entropy (BCE) as the loss function for MTC and the Cross Entropy as the loss function for MLM. BCE loss can be written as follows:
\begin{equation}
	\begin{split}
		\mathcal L_{mtc} = \frac{1}{|L|}\sum_{i=1}^{|L|}(y_{ti} log(\sigma(y_{pi}))+ \\
		(1-y_{ti})log(1-\sigma(y_{pi}))
	\end{split}
\end{equation}
where $\sigma$ denotes sigmoid activation function, $y_t$ is the ground truth label and $y_p$ denotes the predicted results. The final joint loss function is:
\begin{equation}
	\label{eq:loss}
	\mathcal L = \mathcal L_{mtc} + \lambda \mathcal L_{mlm}
\end{equation}

\textbf{Inference}. When inferring, mask all the labels and calculate the probability of all the masked labels. As with training, we express the output of labels as $O_l$ and use the logistic sigmoid function for probability normalization:
\begin{equation}
	\label{eq:p}
	P_l = \sigma(O_l)
\end{equation}
Then, all probability values greater than 0.5 are predicted to be positive labels, otherwise predicted to be negative labels.

\renewcommand{\algorithmicrequire}{\textbf{Input:}} 
\renewcommand{\algorithmicensure}{\textbf{Output:}}
\begin{algorithm}[htb]
	\setstretch{1.1} 
	\caption{Label Mask Multi-label Text Classification}
	\label{alg:Framwork}
	\textbf{Input:} The original sentence $X$ corresponds to the label $Y_t$. \\
	\textbf{Output:} Predicted labels $Y_p$ and a trained model $M$.
	\begin{algorithmic}[1]
		\IF{train}
		\STATE generate templates $\textit{T}$ as Eq~\ref{eq:t} and input tokens with \textit{masks} $X'$ as Eq~\ref{eq:x}.
		\STATE train $M$ by optimizing the joint loss function as Eq~\ref{eq:loss}.
		\ELSE  
		\STATE generate templates $\textit{T}$ by masking all the labels and input tokens $X'$ as Eq~\ref{eq:x}.
		\STATE calculate output $O_l$ by feeding $X'$ into $M$.
		\STATE predict $Y_p$ by probability distribution as Eq~\ref{eq:p}.
		\ENDIF 
	\end{algorithmic} 
\end{algorithm}

\section{Experiments}

\subsection{Datasets}
In view of the wide application of multi-label text classification, we applied our method on different types of data sets to verify the effectiveness of LM-MTC. The statistics of the dataset are shown in Table~\ref{tab:data}. 

\begin{table*}[t]
	\centering
	\begin{tabular}{cccc}
		\hline
		Dataset & Label Sets & Train Size & Test Size \\ \hline
		GAIC                                            & 17         & 24000      & 3000      \\
		AAPD                                            & 54         & 53840      & 1000      \\
		Reuters-21578           & 90         & 8630       & 2158      \\
		RMSC                    & 22         & 5020       & 646       \\
		Emotion                 & 28         & 43410      & 5427      \\
		Toxic                   & 6          & 11357      & 4868    \\ \hline
	\end{tabular}
	\caption{Datasets statistics.}
	\label{tab:data}
\end{table*}

\begin{itemize}
	\item GAIC\footnote{https://tianchi.aliyun.com/competition/entrance/531852}. A competition dataset of desensitized medical texts which all words are replaced by numbers. We combined the dataset of the preliminary match and the semifinal match together, removed the 12 labels added in the semifinal match, and finally constituted a 17-label classified dataset of 30,000 samples. Then, we divide the training set, development set and test set according to the ratio of 8:1:1. We pre-trained a Bert model based on desensitization data and used the parameters of this Bert in the MTC task. 
	
	\item AAPD~\cite{yang2018sgm}. A widely used large-scale classification dataset for multidisciplinary academic papers. The goal is to predict the subject by abstracts.
	
	\item Reuters-21578~\cite{debole2005an}. Reuters news text dataset created in 1987 which has been a standard benchmark for MTC. We follow the classification criteria of ~\cite{ankit2020multi-label} and use 90 categories.
	
	\item RMSC ~\cite{zhao2018review-driven}. Collected from a Chinese popular music website\footnote{https://music.douban.com}. The goal is to distinguish musical styles based on different reviews. 22 styles are defined. We divide the dataset into training/validation/test sets as described in the original paper. Note that since it is a Chinese dataset, we use the Chinese pre-trained Bert model\footnote{https://huggingface.co/bert-base-chinese}. In the process of data preprocessing, we remove non-Chinese and non-English special symbols.
	
	\item Emotion~\cite{dorottya2020goemotions}. A largest manually annotated dataset of 58k English Reddit comments for fine-grained sentiment classification labeled for 27 emotion categories and neutral, 28 categories totally. 
	
	\item Toxic Comments\footnote{https://www.kaggle.com/c/jigsaw-toxic-comment-classification-challenge}. A dataset from Toxic Comment Classification Challenge Competition contains text that may be considered profane, vulgar, or offensive. We remove comments that don't carry any negative sentiment and keep only 16,225 tagged records as our dataset. We split the train/test set randomly in a 7:3 ratio.
\end{itemize}

\subsection{Evalution Metrics}
The same as~\cite{tsoumakas2007random,yang2018sgm}, Hamming Loss and micro-F1 Score are used for the main evaluation metrics. Besides, we also use Accuracy and Micro-Jaccard for further evaluation.
\begin{itemize}
	\item Accuracy. The strict accuracy of multi-label classification. For the given prediction result $Y_{p}\in \mathbf{R}^{|\Gamma|*|L|}$ and ground truth $Y_{t}\in \mathbf{R}^{|\Gamma|*|L|}$, Accuracy is calculated as follows:
	\begin{equation}
		Accuracy = \sum_{i}^{|\Gamma|}{\frac{\Xi(Y_{ti},Y_{pi})}{|\Gamma|}}
	\end{equation}
	where $|\Gamma|$ is the test set size, $\Xi(\cdot)$ is an indicator function. If the corresponding elements at all positions are equal in $Y_{ti}$ and $Y_{pi}$, $\Xi(Y_{ti},Y_{pi})=1$, else $\Xi(Y_{ti},Y_{pi})=0$.
	
	\item Micro-F1~\cite{yang2018sgm}. It can be interpreted as a weighted average of accuracy and recall. It calculates indicators globally by calculating total true positives, false negatives and false positives. 
	
	\item Micro-Jaccard~\cite{baldi2000assessing}. The Jaccard similarity coefficient is defined as the size of the intersection divided by the size of the union of two label sets. 
	
	\item Hamming loss (HL)~\cite{schapire1998improved}. It directly calculates the proportion of misclassified labels. A value of 0 means that all labels for each sample have been assigned the correct label.
\end{itemize}

\subsection{Baselines}
We compare LM-MTC with the widely available baselines:
\begin{itemize}
	\item Binary Relevance (BR)~\cite{boutell2004learning}. Categorize each label separately, regardless of the correlation between the labels. 
	
	\item Classifier Chains (CC)~\cite{read2011classifier}. Consider the high order correlation between labels and convert it into a binary classification chain.
	
	\item CNN~\cite{kim2014convolutional}. Convolutional neural network is used to extract text features and outputs the distribution of labels in the label space.
	
	\item CNN-RNN~\cite{chen2017ensemble}. CNN and RNN are combined for local and global text modeling.
	
	\item Hierarchical Attention Network (HAN)~\cite{yang2016hierarchical}. Model text with hierarchical attention to words and sentences. 
	
	\item HAN+LG~\cite{zhao2018review-driven}. Introduce Label Graph (LG) on the basis of HAN, and the soft association relationship between labels is trained by LG.
	
	\item SGM~\cite{yang2018sgm}. View the MCT as a sequence generation
	problem, and apply a sequence generation model with a novel decoder structure to solve it.
	
	\item Bert~\cite{devlin2018bert}. Self-attention based pretrained language model. Make different fine-tuning for different downstream tasks.
	
	\item Bert+MLM. On the basic Bert classification, additional MLM tasks are added.
	
	\item MEGNET~\cite{ankit2020multi-label}. A graph attention network-based model to capture the attentive dependency structure among the labels.
	
	\item Label-Wise (LW) LSTM with PT and FT~\cite{liu2020label-wise}. A document representation with label-aware information is obtained through a pre-training model and fine-tuned for different downstream tasks. PT denotes the pre-training method. FT denotes the fine-tuned method on downstream task.
\end{itemize}

\begin{figure*}[h]
	\centering
	\subfigure[Top11.]{
		\label{fig:vis_a}
		\begin{minipage}[t]{0.5\linewidth}
			\centering
			\includegraphics[width=2.65in]{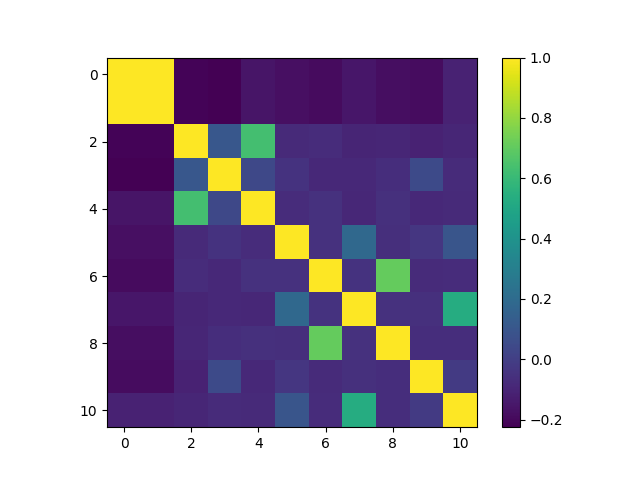}
		\end{minipage}%
	}%
	\subfigure[2nd attention.]{
		\label{fig:vis_b}
		\begin{minipage}[t]{0.5\linewidth}
			\centering
			\includegraphics[width=2.65in]{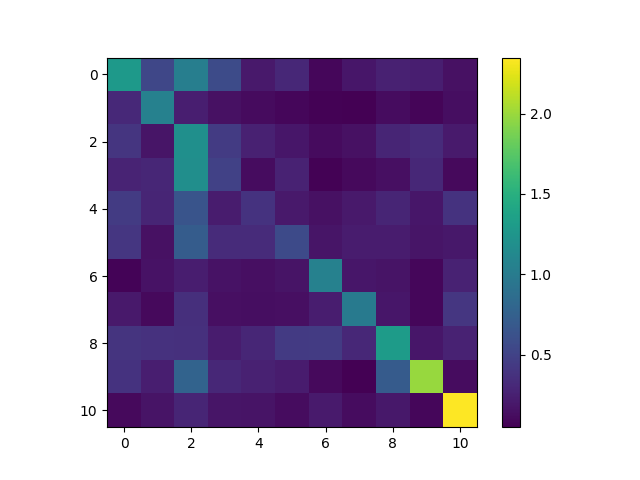}
		\end{minipage}%
	}%

	\subfigure[6th attention.]{
		\label{fig:vis_c}
		\begin{minipage}[t]{0.5\linewidth}
			\centering
			\includegraphics[width=2.65in]{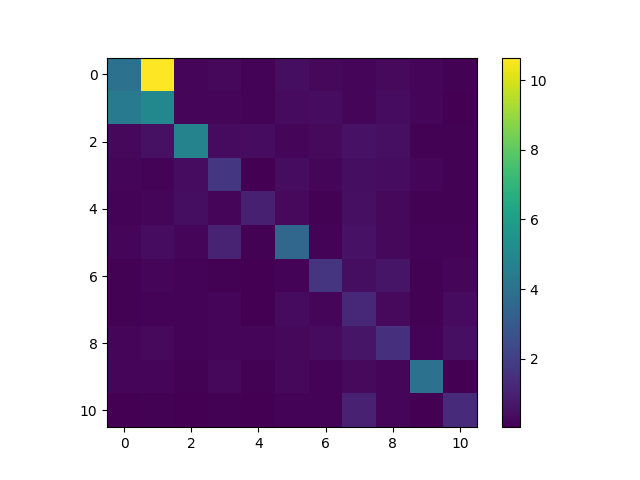}
		\end{minipage}
	}%
	\subfigure[12th attention.]{
		\label{fig:vis_d}
		\begin{minipage}[t]{0.5\linewidth}
			\centering
			\includegraphics[width=2.65in]{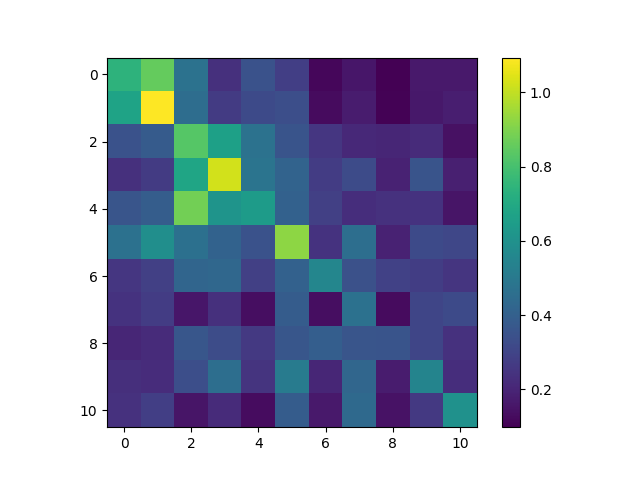}
		\end{minipage}
	}%
	\centering
	\caption{Spearman correlation coefficient of top11 labels and attention visualization of different Bert layers of AAPD test set.}
\end{figure*}

\begin{table*}[t]
	\centering
	\begin{spacing}{1}
		\begin{tabular}{cccccc}
			\hline
			Dataset &Model & Accuracy(+,\%) & Micro-F1(+,\%) & Micro-Jaccard(+,\%) & HL(-)  \\ \hline
			& Bert & 89 & 92 & 86 & 0.02 \\
			& Bert+MLM & 87 & 91 & 84 & 0.02 \\
			& LM  & 9 & 93  & 86  & 0.02 \\ 
			\multirow{-4}{*}{\# GAIC}  & LM+MLM & \textbf{92} & \textbf{95} & \textbf{9} & \textbf{0.01} \\ \hline
			& Bert & 42.2          & 72.57          & 56.95          & 0.0226          \\ 
			& Bert+MLM  & 40.5&	72.8&	57.23&	0.0227\\ 
			& ML & 41.3 & \textbf{73.73} & \textbf{58.39} & 0.0228         \\ 
			\multirow{-4}{*}{\# AAPD} & LM+MLM   & \textbf{42.3} & 73.61  & 58.23 & \textbf{0.0224} \\ \hline
			& Bert & 85.25& 89.09 & 80.33   & 0.0029   \\ 
			& Bert+MLM & 85.03	&89.4&	80.83&	0.00274   \\ 
			& LM     & 84.15   & 89.6   & 81.15 & 0.0028    \\ 
			\multirow{-4}{*}{\# Reuters-21578} & LM+MLM   & \textbf{85.59} & \textbf{89.97} & \textbf{81.77} & \textbf{0.00264} \\  \hline
			& Bert & 37.46 & 72.21 & 56.51  & 0.0507 \\
			& Bert+MLM   &\textbf{39.63}&	73.18&	57.71&	\textbf{0.0499}\\ 
			& LM  & 36.53 & 72.81 & 57.24 & \textbf{0.0499} \\ 
			\multirow{-4}{*}{\# RMSC}& LM+MLM   & 38.39 & \textbf{73.8} & \textbf{58.47} & 0.0517 \\ \hline
			& Bert & 47.08 & 57.55 & 40.4  & 0.031 \\
			& Bert+MLM   &43.97	&54.7&	37.7&	0.033\\  
			& LM   & 45.35 & 58.02  & 40.86  & 0.0306 \\ 
			\multirow{-4}{*}{\# Emotion}& LM+MLM   & \textbf{47.63}  & \textbf{59.07} & \textbf{41.92} & \textbf{0.0304}  \\ \hline
			& Bert & 52.67  & 85.76 & 75.07 & 0.1043  \\  
			& Bert+MLM   &51.97&	84.96&	73.86&	0.1087\\ 
			& LM & 52.71& \textbf{85.86} & \textbf{75.23} & 0.1048  \\ 
			\multirow{-4}{*}{\# Toxic Comments} & LM+MLM  & \textbf{53.53} & 85.77 & 75.08   & \textbf{0.1041} \\ \hline    
		\end{tabular}
	\end{spacing}
	\caption{Performance over different datasets. The bold represents the optimal results. }
	\label{tab:results1}
\end{table*}

\begin{table*}[t]
	\begin{spacing}{1}
		\begin{tabular}{ccccccc}
			\hline
			& \multicolumn{2}{c}{AAPD} & \multicolumn{2}{c}{Reuters-21578} & \multicolumn{2}{c}{RMSC} \\ \cline{2-7}
			\multirow{-2}{*}{Model} & Micro-F1(+,\%) & HL(-) & Micro-F1(+,\%) & HL(-)  & Micro-F1(+) & HL(-)  \\ \hline
			BR   & 64.6   & 0.0316 & 87.8  & 0.0316  & 41.8  & 0.083  \\
			CC  & 65.4  & 0.0306& 87.9 & 0.0306  & 44.3 & 0.107 \\
			CNN& 65  & 0.0264  & 86.3   & 0.0287  & 59.1 & 0.0702 \\
			CNN-RNN  & 66.4 & 0.0278   & 85.5  & 0.0282  & -    & -   \\
			HAN  &70.81&0.0236&-&-&66.75&0.059 \\
			HAN+LG  & 71.19&0.0235&-&-&68.21& 0.058\\
			SGM   & 71  & 0.0245 & -    & -  & -  & - \\
			Bert  & 72.57 & 0.0226 & 89.09 & 0.0029 & 72.21& 0.0507  \\
			Bert+MLM  & 72.23&0.0231& 89.4 & 0.00274 &73.18& \textbf{0.0499}\\
			MEGNET  & 69.6  & 0.0252  & 89.9  & 0.0252 & -  & -\\ \hline
			LW-LSTM+PT & 71.21  & 0.0239 & -  & -  & 67.04  & 0.0583  \\
			LW-LSTM+FT  & 71.31  & 0.0241 & -  & - & 72.18& 0.0537  \\ \hline
			LM   & \textbf{73.73}  & 0.0228  & 89.6 & 0.0028& 72.81  & \textbf{0.0499} \\
			LM+MLM  & 73.61 & \textbf{0.0224} & \textbf{89.97} & \textbf{0.00264} & \textbf{73.8} & 0.0517 \\
			\hline        
		\end{tabular}
	\end{spacing}
	\caption{Comparison with other baseline methods. The bold represents the optimal results. }
	\label{tab:results2}
\end{table*}

\subsection{Details}
We set the learning rate as 5e-5, the batch size as 16, and the running epoch as 40. We set the warm up epoches ratio to 0.1, set the mask probability of MLM to 0.15. We use AdamW as the optimizer~\cite{loshchilov2018fixing}. All of the code is written using PyTorch and runs on NVIDIA RTX 3090.

\subsection{Overall Results}
We report the experimental results of our method on all datasets in Table~\ref{tab:results1} and compare them with Bert and Bert+MLM. We calculate the Accuracy, Micro-F1, Micro-Jaccard and Hamming-Loss. Compared with BERT, in most cases, LM has significant performance improvement in all evaluation metrics, which indicates that the model performance can be effectively improved after transforming MTC into template populated tasks. We also notice that adding MLM can further improve the performance of ML and Bert, which illustrates the effectiveness of joint training. 

We explain this phenomenon from the essence of Bert. When we input the label and the original sentence together into Bert, it is equivalent to constructing the context for the label, and self-attention can sensitively capture such context relations that do not exist in the original sentence. In this way, we introduce the association among labels which can increase the model's ability to understand the label context. In addition, since Bert is essentially an MLM, allowing Bert to continue learning the mask for different downstream tasks can improve its performance. 

We also compared LM-MTC with some other commonly used baseline models. The results are shown in Table~\ref{tab:results2}. We notice that approaches with labels relationships is often superior to methods that do not consider the relationship between labels. For example, CC is better than BR, and HAN+LG is better than HAN. This demonstrates the necessary of label relevance. We also notice that our method outperforms all baselines, because our approach takes advantage of Bert's powerful context-capture capabilities to capture semantic associations between labels. This approach is more efficient than some methods of explicit label association modeling, such as graph-based or chain-based methods. 

\subsection{Analysis}

\subsubsection{Attention Visualization}

The middle layer of Bert has been shown to adequately capture semantic relationships between words~\cite{jawahar2019what}. In LM-MTC, each potential label can be treated as a word, so we verify how the LM-MTC captures the correlation of labels by visualizing the attention of each layer. 

Figure~\ref{fig:vis_a} shows the Spearman correlation between different labels of AAPD testset. To facilitate the observation, we chose the top11 labels with high relevance. Similar to Figure~\ref{fig:relation}, test labels have a similar correlation distribution with train dataset. 

After that, we take the attention output parameters of different Bert layers. We average all the attention heads and select the attention scores between all the label pairs. We add up all the batch data to get a global score matrix of attention on testset. We select the attention matrix of the 2nd layer (Figure~\ref{fig:vis_b}), the 6th layer (Figure~\ref{fig:vis_c}) and the last layer (Figure~\ref{fig:vis_d}) for visualization. Bert in the shallow layer learns some rough information, the 6th layer pays more attention to the local correlation, and the attention of the last layer is closer to the original label correlation distribution. This shows that deep Bert can capture the correlation between labels, which also provides a valid explanation for the advantages of LM-MTC.

\subsubsection{MLM Loss Ratio}
\begin{figure}[h]
	\centering
	\includegraphics[scale=0.5]{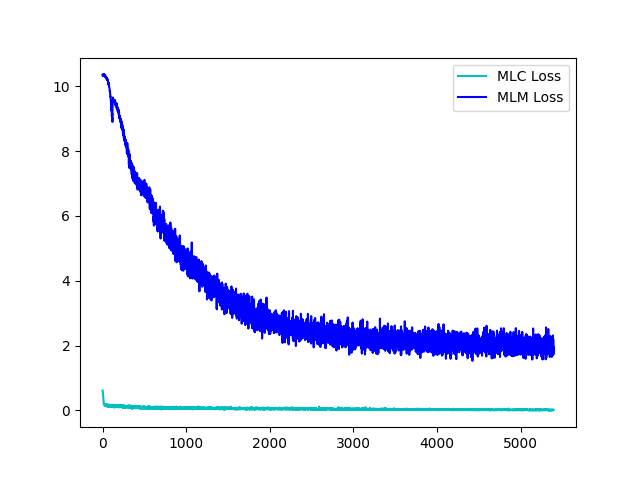}
	\caption{Changes of every 10 batch loss function during training.}
	\label{fig:loss_plot}
\end{figure}

For multi-tasking learning, different task Loss should be of similar magnitude~\cite{chen2018gradnorm}. According to the changes in MTC and MLM loss as shown in Figure~\ref{fig:loss_plot}, we select different $\lambda$ to investigate the effect of MLM task weight on model performance. 

\begin{figure}[htbp]
	\centering
	\subfigure[RMSC.]{
		\label{fig:rmsc_bar}
		\begin{minipage}[t]{0.5\linewidth}
			\centering
			\includegraphics[width=2.65in]{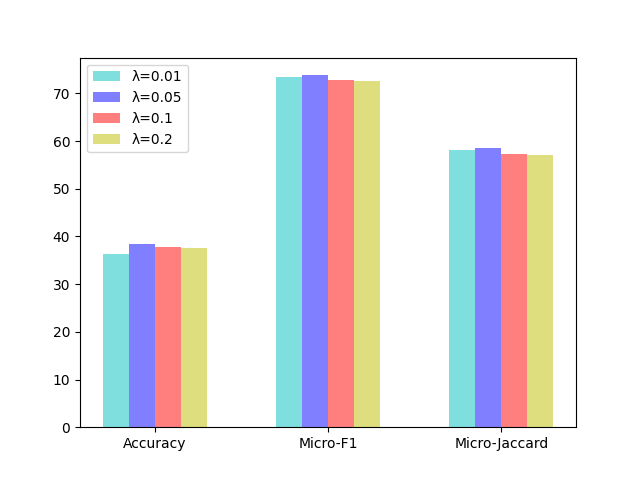}
		\end{minipage}%
	}%
	\subfigure[Emotion.]{
		\label{fig:_bar}
		\begin{minipage}[t]{0.5\linewidth}
			\centering
			\includegraphics[width=2.65in]{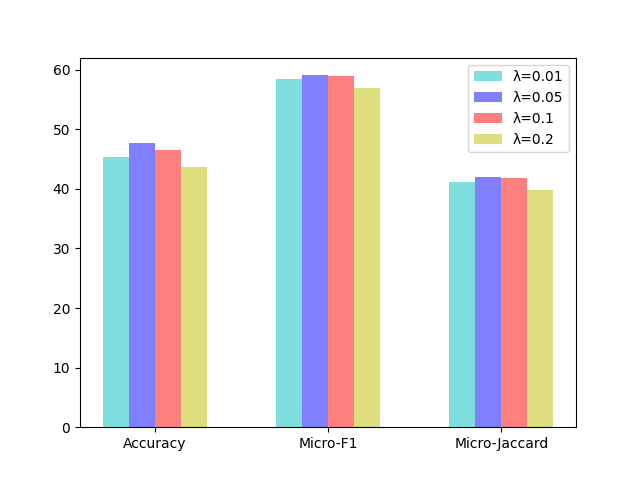}
		\end{minipage}%
	}%
	\centering
	\caption{Model performance under different $\lambda$.}
\end{figure}

As is shown in Figure~\ref{fig:_bar}, for two different datasets, RMSC and Emotion, the best performance occurs at $\lambda=0.05$, $\lambda$ is too large or too small will have a negative impact on the model. Therefore, So to keep the MTC task dominant, and to make the MLM task have enough impact, we set $\lambda=0.05$ for all datasets. 

\subsubsection{Different Mask Strategies}
In fact, different templates may have different effects on the same task~\cite{shin2020autoprompt}. As a comparison, we assign the same token to each label in different positions, that is, we modify the template in Eq~\ref{eq:t} as follows:
\begin{equation}
	\label{eq:t_same}
	\begin{split}
		\color[RGB]{0,0,0}{[LS][YES][LE]} \\
		\color[RGB]{0,0,0}{[LS][NO][LE]} \\
		\color[RGB]{0,0,0}{[LS][MASK][LE]}
	\end{split}
\end{equation}

\begin{table*}[t]
	\centering
	\begin{tabular}{cllcl}
		\hline
		& \multicolumn{2}{c}{RMSC} & \multicolumn{2}{c}{Emotion}       \\
		\multicolumn{1}{l}{}                            & Diff        & Same       & \multicolumn{1}{l}{Diff} & Same   \\ \hline
		Accuracy                                        & 38.39       & 21.67      & 47.63                    & 37.31  \\
		Micro-F1                                        & 73.8        & 60.63      & 59.07                    & 51.38  \\
		Micro-Jaccard           & 58.47       & 43.5       & 41.92                    & 34.57  \\
		HL                      & 0.0517      & 0.065      & 0.0304                   & 0.0323 \\ \hline
	\end{tabular}
	\caption{The influence of different mask strategies on the results.}
	\label{tab:strategies}
\end{table*}
We repeat the experiment on RMSC and Emotion, and the results are shown in Table~\ref{tab:strategies}. When all labels use the same templates, there is a significant performance drop. This means that different label templates can better provide identification information for Bert, but same template does not. 

\section{Conclusion}
In this paper, we propose LM-MTC model for multi-label text classification. We build prefix templates for multiple labels, transform MTC into cloze task, and combine training with MLM to improve the performance of the model under a variety of evaluation metrics. Further, we explain the ability of LM-MTC to capture the potential association between labels and LM-MTC can perform well in tests against multiple types of datasets. In future work, we will explore the ability of different templates to combine with different pre-trained language models.

\section{Acknowledgements}
This work was supported by National Natural Science Foundation of China
(NSFC), "From Learning Outcome to Proactive Learning: Towards a Humancentered AI Based Approach to Intervention on Learning Motivation" (No.62077027).

\bibliography{ref}

\end{document}